\documentclass{article}
\usepackage{graphicx}
\usepackage{hyperref}
\usepackage{microtype}
\usepackage{subfigure}
\usepackage{booktabs}
\usepackage{array}
\usepackage{amsmath}
\usepackage{amssymb}
\usepackage{mathtools}
\usepackage{amsthm}
\usepackage{times}
\usepackage{latexsym,multirow}
\usepackage{textcomp}
\usepackage[numbers,sort]{natbib} 
\usepackage{enumitem}
\usepackage{algorithm}
\usepackage{algpseudocode}
\DeclareMathOperator*{\argmax}{arg\,max}

\usepackage[capitalize,noabbrev]{cleveref}
\usepackage{tcolorbox}
\usepackage[a4paper, margin=1.5in]{geometry}

\title{Policy Learning with a Natural Language Action Space: A Causal Approach}
\author{
    \makebox[\textwidth][c]{Bohan Zhang, Yixin Wang, Paramveer S. Dhillon}\\
    University of Michigan \\
    \texttt{zbohan@umich.edu}, \texttt{yixinw@umich.edu}, \texttt{dhillonp@umich.edu}
}
\date{}
\begin{document}

\maketitle

\begin{abstract}
This paper introduces a novel causal framework for multi-stage decision-making in natural language action spaces where outcomes are only observed after a sequence of actions. While recent approaches like Proximal Policy Optimization (PPO) can handle such delayed-reward settings in high-dimensional action spaces, they typically require multiple models (policy, value, and reward) and substantial training data. Our approach employs Q-learning to estimate Dynamic Treatment Regimes (DTR) through a single model, enabling data-efficient policy learning via gradient ascent on language embeddings. A key technical contribution of our approach is a decoding strategy that translates optimized embeddings back into coherent natural language. We evaluate our approach on mental health intervention, hate speech countering, and sentiment transfer tasks, demonstrating significant improvements over competitive baselines across multiple metrics. Notably, our method achieves superior transfer strength while maintaining content preservation and fluency, as validated through human evaluation. Our work provides a practical foundation for learning optimal policies in complex language tasks where training data is limited.
\end{abstract}

\section{Introduction}
\begin{figure*}[h]
\begin{small}
    \centering
    \includegraphics[width=.65\textwidth]{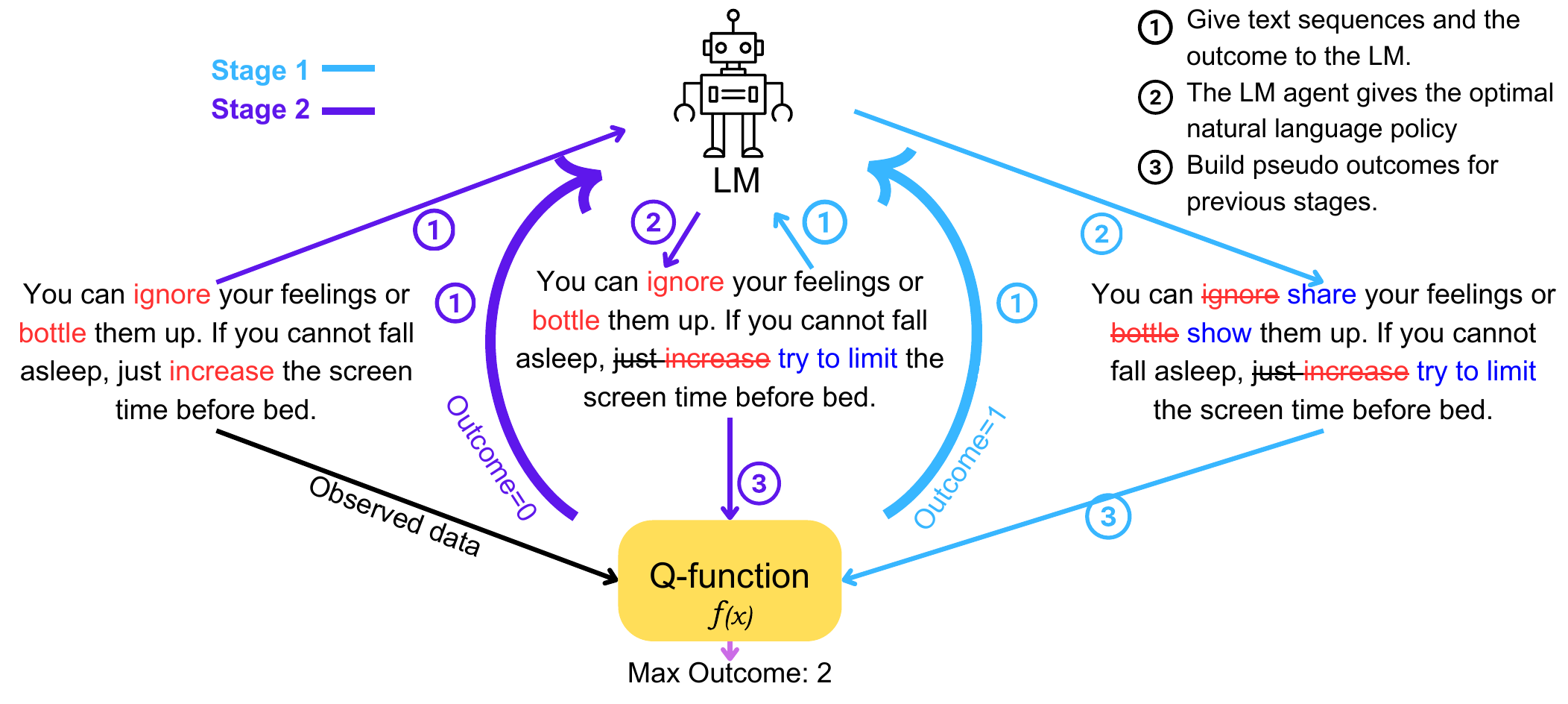}
    \caption{\small{Multi-stage decision-making with a natural language action space. For simplicity, the figure shows a two-stage example. The task is to transfer a sequence of textual interventions for mental health issues from ineffective to effective. The texts with ineffective signals are highlighted in red and the optimal natural language policy given by the LM are highlighted in blue. Steps within the same stage of Q-Learning are represented using the same colored arrows.}}
    \label{fig:gen-Q}
    \end{small}
\end{figure*}

Large Language Models (LLMs) have revolutionized artificial intelligence with their remarkable ability to understand and generate human-like text across various applications \cite{brown2020language,achiam2023gpt}. However, harnessing their full potential in multi-stage decision-making scenarios, especially in human-AI collaboration, remains a significant challenge. Consider a mental health treatment scenario where an LLM assists psychiatrists in developing personalized therapeutic interventions. The psychiatrist initially outlines a treatment approach based on the patient's symptoms and history. The LLM then analyzes this input to suggest refinements that align with evidence-based practices and therapeutic guidelines. The psychiatrist reviews and builds upon these suggestions, incorporating their clinical expertise, and this collaborative cycle continues until an optimal intervention strategy emerges. This real-world application exemplifies the challenges of multi-stage decision-making in high-stakes medical contexts, where each action must navigate an expansive natural language space while maintaining clinical validity and effectiveness. The fundamental challenge lies not just in the high-dimensional, continuous nature of language, but in making sequential decisions that both preserve critical medical information and optimize therapeutic impact~\cite{he2015deep,jaques2019way}. Each refinement must consider both immediate improvements and their influence on subsequent treatment decisions, creating a complex optimization problem that extends beyond simple text enhancement. Each decision not only affects the immediate outcome but also influences all subsequent steps and final results, creating a complex interdependence that is difficult to optimize~\cite{sutton1998reinforcement,silver2014deterministic}.

Multi-stage decision-making is fundamental in numerous domains where sequential actions significantly influence future outcomes, such as dialogue systems, interactive machine translation, and text summarization~\cite{wen2016network,peng2017composite,zhang2019budgeted,dong2018banditsum,gu2021memsum,lam2019interactive,huang2021transmart}. To address these challenges, researchers have traditionally employed policy learning approaches, aiming to optimize decision-making strategies through experience and feedback~\cite{sutton1998reinforcement,silver2014deterministic}. However, many existing techniques primarily focus on discrete action spaces~\cite{mnih2015human,lillicrap2015continuous}, such as predefined dialogue acts in conversational AI~\cite{kwan2023survey,chen2017survey}. While effective in certain scenarios, these discrete representations limit the ability to directly optimize natural language actions. The natural language action space presents unique challenges due to its unstructured nature, high dimensionality, and vast combinatorial possibilities~\cite{he2015deep,jaques2019way}. This complexity makes traditional optimization and policy learning approaches less effective, as they struggle to efficiently explore and exploit the richness of natural language.

Further, identifying the optimal policy in multi-stage decision-making requires a causal inference approach, particularly when dealing with natural language actions and delayed rewards (the therapeutic impact of the psychiatric intervention only observed at the end). Recent advances in LM alignment through reinforcement learning, such as Proximal Policy Optimization (PPO)-based approaches~\cite{InstructGPT,ConstitutionalAI}, have shown promise for such setups but typically require learning multiple models (policy, value, and reward) and substantial training data for stable optimization. The key challenge lies in efficiently estimating optimal policies while addressing confounding between LM actions and user inputs. To address this, we turn to Q-learning and Dynamic Treatment Regimes (DTR) \cite{watkins1989learning,murphy2003optimal}, which provide a data-efficient framework for optimizing sequences of decisions through a single model. These methods allow us to frame the problem in terms of potential outcomes, enabling the estimation of causal effects while addressing time-varying confounding \cite{robins2004optimal,hernan2020causal}.

Building on this foundation, we propose a novel causal framework that integrates a language model with a task classifier functioning as a Q-function. Our approach (summarized in Figure~\ref{fig:gen-Q}) employs gradient ascent on text embeddings to optimize the classifier's output, effectively maximizing outcomes over multiple stages. This enables efficient policy learning in natural language action spaces while avoiding direct text manipulation (in the high-dimensional space). At the core of our approach is a novel decoding strategy that translates optimized embeddings back into natural language. 
This data-efficient framework allows us to learn effective policies for multi-stage language tasks from limited data, providing a robust foundation for structured text transformation tasks that require sequential refinement.

We evaluate our framework on three distinct tasks: mental health intervention refinement, hate speech countering, and sentiment style transfer. Our experimental results demonstrate consistent improvements over state-of-the-art baselines across multiple metrics, with particularly strong performance in outcome transfer strength - surpassing previous approaches by up to 30\% on key benchmarks. Human evaluations corroborate these findings, showing that our approach achieves more balanced performance across fluency, content preservation, and transfer strength, leading to significantly higher success rates in multi-stage text transformation tasks.

Our key contributions can be summarized as follows:

\begin{itemize}[leftmargin=*]
    \item We introduce a novel causal framework for policy learning in sequential decision-making problems with a natural language action space.
    \item We propose a new algorithm for policy learning with natural language actions using gradient ascent on text embeddings, suitable for multi-stage decision problems.
    \item We develop a novel decoding strategy that translates optimized embeddings back into natural language.
\end{itemize}

\section{Related Work}
Our work is related to three strands of prior work:

\paragraph{Policy Learning for Multi-Stage Decision-Making: } 
Policy learning for multi-stage decision-making has been widely explored in healthcare~\cite{moodie2012q,lei2012smart} and personalized medicine~\cite{cain2010start,wahed2004optimal} to optimize treatment strategies over time. This approach has also been applied to dialogue systems~\cite{peng2017composite,zhang2019budgeted}, interactive machine translation~\cite{lam2019interactive,huang2021transmart}, and extractive text summarization~\cite{dong2018banditsum,gu2021memsum}. While these studies offer valuable insights into multi-stage decision-making, they typically deal with discrete actions and do not address scenarios involving high-dimensional text actions.

\paragraph{Reinforcement Learning for a Natural Language Action Space: }
He et al.~\cite{he2015deep} proposed an architecture for handling natural language action spaces in text-based games, using separate neural networks to embed state and action text before combining them to approximate the Q-function. Wang et al.~\cite{wang2024language} introduced a method that dynamically adapts the prior of a pre-trained language model using mutual information regularization to implicitly reduce the action space. Our setting differs from these approaches in that it deals with continuous rather than discrete action spaces, focuses on offline learning, and lacks explicit state transitions.

\paragraph{Causal Inference for Text: } 
Our work also relates to the emerging field of causal inference for language tasks. Veitch et al.~\cite{veitch2020adapting} pioneered the use of language model embeddings and topic modeling to mitigate textual confounding in treatment effect estimation. This approach has been extended by researchers like Egami et al.~\cite{egami2022make}, Pryzant et al.~\cite{pryzant2020causal}, and Imai and Nakamura~\cite{imai2024causal}, who focus on leveraging latent representations of high-dimensional text as confounders or treatments. These studies highlight the importance of extracting low-dimensional latent representations to accurately estimate treatment effects in complex, high-dimensional textual data~\cite{louizos2017causal,kim2021counterfactual,zhang2024causal,wang2021desiderata}.

Our work synthesizes elements from these areas, addressing the challenge of finding optimal treatments in a high-dimensional natural language space to maximize outcomes in sequential decision problems.

\section{Policy Learning with a Natural Language Action Space}

We frame the policy learning problem in multi-stage decision-making as
a causal inference problem, and then discuss the policy learning algorithm
with a natural language action space.

\subsection{Policy Learning in Multi-Stage Decision-Making}
As a running example of multi-stage decision-making, we consider a mental health intervention setting where a language model (LM) assists psychiatrists in developing treatment strategies. The process operates through alternating steps: the psychiatrist initiates by writing therapeutic interventions based on the patient's mental health status, which the LM then refines to enhance clinical effectiveness. The psychiatrist subsequently builds upon these refinements, and this cycle continues. After several iterations, the final intervention strategy, jointly developed by the psychiatrist and LM, is evaluated based on its therapeutic effectiveness with patients.

Our dataset comprises the psychiatrist's initial interventions, the LM's refinements at each stage, and the final intervention's effectiveness assessment. The goal is to identify the optimal policy for the LM: determining how it should refine clinical text to maximize therapeutic impact (only observed at the end) while working collaboratively with mental health professionals.

\textbf{Multi-stage decision-making as a causal inference:} Identifying the optimal policy in multi-stage decision-making requires causal inference. Historical data can be misleading: an LM's actions associated with the best final outcomes may not be inherently effective. Rather, these actions might appear seemingly effective only because it coincides with good user inputs. Thus the key challenge in policy learning lies in finding LM actions that have a large \emph{causal effect} on the outcome. 

To formalize the causal problem, we denote the text to be modified at time $t$ (e.g., the user input) as $L_t$ with $t=1,..., T$. At each time stage, the LM executes a refinement action $A_t$ (such as the post-refinement text) to modify the target text with certain goals. The sequence of LM's actions $\bar{A}_t=A_{1:t}$ is the  ``treatment.'' The \emph{potential outcome} $Y_T(a_1,...,a_T)$ is the potential reward if the series of actions $(a_1,...,a_T)$ were applied~\cite{imbens2015causal,pearl2009causality,hernan2010causal}.
For example, the potential outcome can be whether the modified medical intervention is effective or not. For notational convenience, we define $\bar{L}_t = L_{1:t}$ and $H_t=(\bar{L}_t,\bar{A}_{t-1})$ which includes all historical variables for $A_t$. The goal is to find a sequence of actions $\bar{a}_T$ that maximize the final potential outcome $Y_T(a_1,...,a_T)$, also known as identifying the optimal dynamic treatment regimes~\citep{watkins1989learning,murphy2005generalization,sutton2018reinforcement}.

\textbf{Q-learning for estimating optimal dynamic treatment rules:} One algorithm to construct the optimal sequence of actions is Q-learning~\citep{watkins1989learning,murphy2005generalization,sutton2018reinforcement}. It identifies causally effective actions by evaluating outcomes as a function of both the current action and all the historical LM actions and user inputs. Technically, Q-learning relies on the optimal Q-function for each stage $t$, defined in a recursive way: for $t < T$,
\begin{align}
\nonumber
&Q_t(H_t, A_t) = \mathbb{E}\Big[ \max_{a_{t+1}} Q_{t+1}\big(H_{t+1}, \bar{a}_{t+1}\big) \,\Big|\, H_t, A_t \Big]\\
\nonumber
&Q_T(H_T, a_T) = \mathbb{E}\big[ Y_T(\bar{a}_T) \,\big|\, H_T, a_T \big].
\end{align}

To find the optimal actions, Q-learning in dynamic treatment regimes considers a \textit{pseudo outcome} at each stage,
\begin{equation}
\nonumber
    \Tilde{Y}_t= \max_{a_{t+1}} Q_{t+1}(H_{t+1},a_{t+1})
\end{equation}
for $t\neq T$. If $t=T$, $\Tilde{Y}_T~=~Q_T(H_T, A_T)~=~Y_T$ ~\citep{murphy2003optimal}. 

If the Q-functions were known, the optimal action at stage $t$ can be found using a backward induction argument as in dynamic programming: given the historical data $H_t=h_t$, the optimal action $a^*_t$ is $a^*_t (h_t)~=~\argmax_{a_t} Q_t(h_t,a_t)$. However, in practice, the true Q-functions are unknown; one must estimate the Q-functions from data. If the outcome is categorical, we can estimate the Q-functions by fitting a neural network text classifier at each stage, since the inputs are text sequences.

\subsection{Policy learning with a Natural Language Action Space}

Classical Q-learning algorithms cannot be applied to finding optimal natural language policies for LMs. These algorithms require searching over all possible actions \( A_t \) and selecting one that maximizes the value of the Q-function. Here the space of actions entails natural language; directly searching over the space of natural language is not feasible.

We address this challenge in two steps. We first consider an encoding of natural language representations and find optimal text-based actions via gradient ascent in the representation space. We then train a specialized language model to decode the optimal text representation into natural language.

\textbf{Q-learning with text representations: } For simplicity of exposition, consider the case of a binary outcome. We approximate the Q-function at each stage $t$ by a binary text classifier $f_t$ that maps the \emph{representations} of historical data and the current action $(h_t, a_t)$ into a probability of positive outcome $y^+$, $P_{f_t}(y^+|\mathbf{a_t},h_t)$, where $y^+$ is the label for the positive outcome, e.g., the effective intervention. In particular, both the historical data $h_t$ and the current text-based action $a_t$ are given as their text representations, following a pre-trained language model encoder, which we will detail in the next paragraph. We then perform Q-learning with the text representations: we maximize the output confidence of the positive outcome from $f_t$ by performing gradient ascent on the high-dimensional representation of natural language actions. The objective function is: 
\begin{equation}
\nonumber
    \mathcal{L}_t(a_t) = - \log (P_{f_t}(y^+|a_t,h_t)).
\end{equation}
We update the embedding representation of $a_t$ iteratively to obtain a higher probability of the positive outcome. We consider the optimal action $a^*_t$ as the action when gradient descent converges.

\begin{figure}[h]
\begin{small}
    \centering
    \includegraphics[width=.48\textwidth]{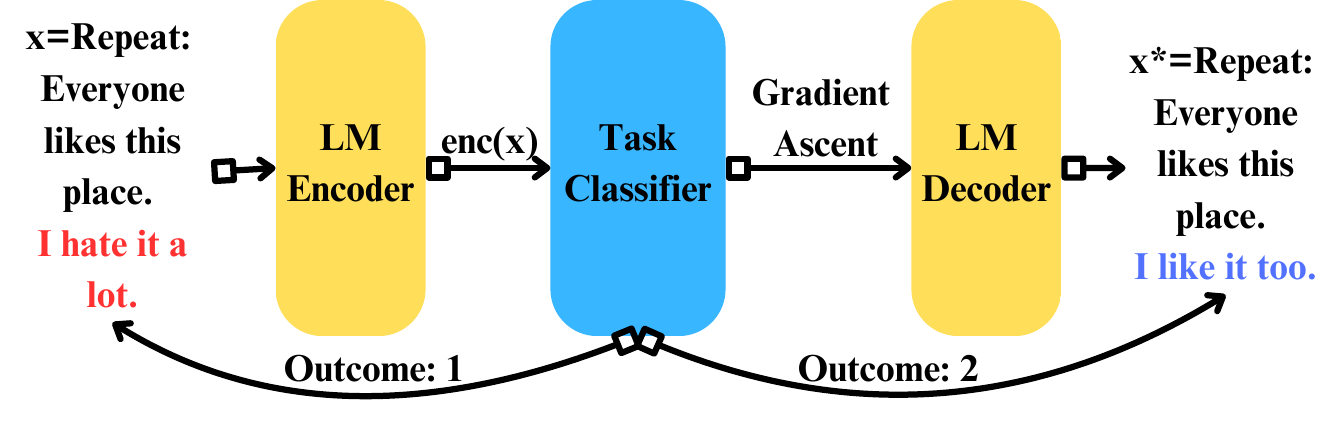}
    \caption{The maximization process of Q-learning in each stage under our framework.}
    \label{fig:gen-max}
    \end{small}
\end{figure}

\begin{algorithm}
\begin{small}
\caption{Policy Learning for Natural Language Action Space {\it (NLPolicyLearn)}}
\begin{algorithmic}
    \State \textbf{Input:} A finetuned T5 following \textit{Repeat} prompt; A dataset of sequences of texts with no editing action initially $\{\bar{L}_{iT}, \bar{A}_{iT}=\bar{L}_{iT},\bar{Y}_{iT}\}_{i=1}^{I}$ 
    \State \textbf{Output:} The optimal sequences of editing actions $\{\bar{A}^*_{iT}\}_{i=1}^I$ 
    \State Let $\{\Tilde{Y}_{iT}\}_{i=1}^I \leftarrow \{Y_{iT}\}_{i=1}^I$
    \For{stage $t$ in $T,...,1$}
        \State 1. Fit a text classifier $f_{t}$ for pseudo outcome $\{\Tilde{Y}_{it}\}_{i=1}^I$ against $\{(H_{it},A_{it})\}_{i=1}^I$ to approximate $Q_t$.
        \State 2. Maximize $f_t$ output by applying gradient ascent on $\{A_{it}\}_{i=1}^I$ .
        \State 3. Decode the $\argmax$ $\{A^*_{it}\}_{i=1}^I$ to get the optimal editing actions.
        \State 4. $\Tilde{Y}_{it-1} \leftarrow Q_t(H_{it},a^*_{it})$ $\forall$ $i$.
    \EndFor
\end{algorithmic}\label{alg:p-learn}
\end{small}
\end{algorithm}

\textbf{Decoding optimal text representations into natural language:} Since Q-learning with text representations only returns the representation of the optimal action, we next develop a novel decoding strategy to decode it into natural language. Specifically, we train a specialized encoder-decoder language model that can take in a prompt ``\textit{Repeat [TEXT]: }'' and output ``\textit{TEXT}.'' The encoder-decoder structure of this language model will first encode the prompt ``\textit{Repeat [TEXT]: }'' into a representation and then decode it to the natural language output ``\textit{TEXT}''; it provides an implicit mapping of the encoder representation of ``\textit{Repeat [TEXT]: }'' to the natural language output ``\textit{TEXT}''.

With this language model, we perform Q-learning in the representation space of its encoder ($enc$). We learn the Q-function by fitting a text classifier as a function of the encoder representation of ``\textit{Repeat:} $(H_t, A_t)$''. We then identify the optimal action as \[\text{enc}(a_t^*) = \argmax_{\text{enc}(a_t)} Q_t \left(\text{enc}\left( \operatorname{Repeat:~} H_t \right),\; \text{enc}(a_t) \right)
\] via gradient ascent. We finally use the specialized language model to decode $enc(a_t^*)$ into the natural language actions $a_t^*$. Specifically, we plug the encoding representations $enc\left( \operatorname{Repeat:~} H_t \right)$ and $enc(a_t^*)$ into the language model to output the text $H_t, a_t^*$, revealing the optimal natural language action $a_t^*$.

We finally note that this decoding algorithm is applicable to both encoder-decoder language models and decoder-only models. While we use a T5 model~\cite{raffel2020exploring} in the empirical studies, we could have also used a decoder-only model: One can perform a maximization process on the output representation of any layer in the language model, and then feed the modified embeddings back into the model to continue decoding.

We summarize the complete algorithm of policy learning for natural language action space in Algorithm~\ref{alg:p-learn}; we also illustrate the search and decoding for the optimal text-based action in Figure~\ref{fig:gen-max}. In practice, we find that the updated embedding can achieve a high probability of the positive label from the text classifier within just a few iterations of gradient ascent. To ensure the fluency of the decoded output, we consider all embeddings from iterations that increase this probability. We then select the embedding that, when decoded into natural language, yields the lowest negative log-likelihood (NLL) as evaluated by GPT-2-Large~\cite{radford2019language}. This selection process balances the objectives of outcome transfer and fluency of natural language.

\section{Experiments and Results}
\begin{table*}[h!]
\centering
\resizebox{0.9\textwidth}{!}{
\begin{tabular}{@{}p{2cm}>{\centering\arraybackslash}p{4.5cm}>{\centering\arraybackslash}p{4.5cm}>{\centering\arraybackslash}p{4.5cm}@{}}
\toprule
\textbf{Dataset} & \textbf{IHS} & \textbf{DIALOCONAN}& \textbf{YELP} \\ \midrule
\textbf{Tasks} & Mental Health Intervention & Hate Speech Countering & Sentiment Style Transfer \\
\textbf{Outcome} & Effectiveness of Intervention & Toxicity of Speech & Positivity of Sentiment \\
\bottomrule
\end{tabular}
}
\caption{Overview of datasets and their associated tasks and outcomes.}
\label{tab:datasets}
\end{table*}
To evaluate the effectiveness of our proposed approach, we conducted extensive experiments on both real-world and synthetic datasets. Our experiments are conducted on three different domains including mental health intervention, hate speech countering, and sentiment style transfer. All of them require estimating optimal natural language policy in multi-stage decision-making settings.

\subsection{Real-World Datasets}

\textbf{Intern Health Study} (\textbf{IHS})~\cite{necamp2020assessing} investigated the effectiveness of real-time moderation strategies in a mobile health intervention for medical interns. The study assessed how different intervention prompts affect mental well-being through medical interns' self-reported responses to the patient health questionnaire (PHQ-9) over a year. The PHQ-9 measures depression severity based on nine diagnostic criteria (sleep, appetite, concentration, etc.) from the Diagnostic and Statistical Manual of Mental Disorders\cite{american2013diagnostic}.

Since the actual textual interventions are not publicly available, we used ChatGPT~\cite{ChatGPT} to generate interventions based on changes in interns' criteria. Following the study's methodology, we identified each participant's most deteriorated criterion every three months and generated corresponding medical interventions. For example, if sleep quality showed the greatest decline among the nine indicators, the model might suggest: ``{\it Use white noise or calming music.}'' We designed the generation process to produce effective interventions (like the example above) with 50\% probability, and ineffective or counterproductive interventions (such as ``{\it Increase screen time before bedtime.}'') otherwise. While these interventions are synthetically generated, they are grounded in real PHQ-9 trajectories and temporal patterns of mental health deterioration, making this dataset particularly suitable for evaluating real-time intervention refinement strategies.

\textbf{DIALOCONAN} ~\cite{bonaldi2022human} contributes the first large-scale dialogue dataset for training multi-turn counter-narrative (CN) generation models against hate speech (HS). The authors use a hybrid human-machine approach to generate the dialogues: concatenating and paraphrasing existing HS/CNpairs and using DialoGPT~\cite{zhang-etal-2020-dialogpt} and T5~\cite{raffel2020exploring} to generate completely new dialogues.~\citet{zhang2024causal} provided a non-hate speech rewrite for each instance of hate speech. Given a sequence of potential hate speech, our model aims to transform them into non-hate speech with less toxicity.

\textbf{Yelp Review Dataset}~\cite{zhang2015character} contains over 200k sentiment-labeled reviews. We manually concatenate a pair of reviews into one sequence of reviews. Our model attempts to rewrite sequences of reviews from negative to positive sentiment.

Our experiments are conducted in a two-stage setup, though our framework is more general and can be extended to more stages. For the training dataset, using IHS as an example, we construct sequences by randomly selecting $x$ pairs for each of the following text combinations: effective-ineffective, ineffective-effective, effective-ineffective, and effective-effective. In total, there will be $4x$ training sequences. The final outcome is the number of effective interventions up to the final stage. In the test dataset, we randomly place $y$ ineffective sentences in one of the two stages. We also implement a one-stage setup for comparison, where all ineffective sentences are placed in the single stage. The DIALOCONAN and Yelp datasets are processed in the same manner. The values of $x$ for IHS, DIALOCONAN, and Yelp are 750, 625, and 900, respectively, while the values of y are 600, 600, and 500.  An overview of the three datasets is shown in Table \ref{tab:datasets}.

\subsection{Baselines and Model Variants}
Our model has two main variants. The first is a two-time sampling (TTS) method addressing gradient ascent plateaus, where we update the random seed and perform gradient ascent twice, selecting the output with a better outcome and smaller edit distance. The second is a one-stage (OS) setup where all sentences with negative outcomes are placed in one stage. Following~\citet{zhang2024causal}, we also implemented a naive baseline without Q-learning, omitting maximization and outcome propagation. We applied Semi-Supervised Variational Autoencoder (SSVAE)~\cite{kingma2014semi} to the encoder representation to capture relevant style features for text reconstruction. 

Although different from the purpose of our study, style transfer methods are generally applicable to these three datasets. We compare our work with two style transfer papers:~\citet{luo-etal-2023-prompt}'s Prompt-based Editing (PE) and~\citet{li2018delete}'s DeleteAndRetrieve (D\&R). More details on training, baseline models, T5 model with~\textit{Repeat} prompt, and gradient ascent process are in Appendix~\ref{app:exp-set}.

\subsection{Evaluation metrics}
Following previous studies~\cite{luo-etal-2023-prompt,he2016ups,suzgun2022prompt}, we evaluated our refinements in three aspects: outcome transfer strength, content preservation, and fluency. We use the following metrics:

\textbf{Content Similarity:} We use SentenceTransformer~\cite{reimers2019sentence} to compute the similarity between the output and the original input~\cite{hallinan2023steer,luo-etal-2023-prompt}, indicating how well the generated output preserves the original content.

\textbf{Transfer Strength (TS):} We train a RoBERTa-base classifier~\cite{liu2019roberta} for each task individually using all single sentences from the training dataset to determine if our model has maximized the outcome. TS is the fraction of outputs across all stages classified as the positive outcome.

\textbf{Fluency:} We measure fluency using the average token-level perplexity of the output, calculated by GPT-2-Large \cite{radford2019language}. Lower scores indicate better fluency. 

We report these metrics for all rewritten negative samples across stages. As the metrics may have different ranges, for a robust comparison, following~\citet{luo-etal-2023-prompt,li2020unsupervised}, we report the \textbf{Geometric Mean} (\textbf{GM}) and \textbf{Harmonic Mean} (\textbf{HM}) of all metrics mentioned above. Since lower fluency values are better, the mean is calculated using $1/\log(\text{fluency})$. The other two metrics are also used as percentage numbers. For the synthetic dataset, which has clearly defined signals, we instead count whether the generated outputs contain the maximum possible number of positive signals.

\subsection{Human Evaluations}
To assess the quality of our model's outputs, we conducted a human evaluation using three native or fluent English-speaking graduate students. Participants rated outputs on three criteria using a 5-point Likert scale: 1) fluency, 2) similarity to the original input (preservation), and 3) sentiment (corresponding to transfer strength). Full questions are listed in Appendix \ref{app:human-eval}. We randomly sampled 50 examples from the YELP test dataset and collected corresponding outputs from our two-stage and two-time sampling model, DeleteAndRetrieve, Prompt-Edit, and human references. Each rater evaluated all four outputs for each example, with the order of samples from different models randomized to ensure fair comparison. Following \cite{li2018delete}, we defined a successful rewrite as one receiving a score of 4 or above on all-three criteria. We calculated the proportion of successful rewrites for each model.

\section{Results and Discussions}
In this section, we present and analyze the results of our experiments, comparing the performance of our proposed methods with that of existing approaches across various datasets and evaluation metrics.

\begin{table*}[htbp]
    \centering
    \begin{scriptsize}
    \renewcommand{\arraystretch}{1} 
    \resizebox{0.75\textwidth}{!}{ 
    \begin{tabular}{c|c|c|c|c|c|c|c}
    \hline
    &   & \multicolumn{3}{c|}{\textbf{Ours}} &  \multicolumn{3}{c}{\textbf{Baselines}} \\ 
    \cline{3-8}
    & \textbf{Metrics}  & \textbf{Base} & \textbf{TTS} & \textbf{OS} & \textbf{PE} & \textbf{D\&R} & \textbf{SSVAE}$^*$ \\ 
    \hline

    \multirow{3}{*}{\textbf{IHS}} & \textbf{Similarity} \textcolor{blue}{$\uparrow$} & \underline{80.7} & 74.0 & \textbf{82.8} & 57.0 & 63.6 & 45.8 \\ 
    & \textbf{Strength} \textcolor{blue}{$\uparrow$} & \underline{41.1} & \textbf{57.3} & 23.7 & 23.1 & 23.3 & 43.8\\ 
    & \textbf{Fluency} \textcolor{red}{$\downarrow$} & 138.0 & 142.5 & \underline{130.7} & 447.4 & \textbf{100.2} & 77.3 \\
    & \textbf{GM} \textcolor{blue}{$\uparrow$} & \underline{40.7} & \textbf{44.1} & 34.3 & 27.8 & 31.8 & 35.9 \\
    & \textbf{HM} \textcolor{blue}{$\uparrow$} & \underline{34.9} & \textbf{37.2} & 29.1 & 24.6 & 28.7 & 34.0 \\ \hline
    
    \multirow{3}{*}{\textbf{Yelp}} & \textbf{Similarity} \textcolor{blue}{$\uparrow$} & 65.3 & 65.3 & \underline{69.2} & \textbf{69.8} & 65.8  & 33.5\\ 
    & \textbf{Strength} \textcolor{blue}{$\uparrow$} & 77.7 & \textbf{90.5} & 72.9 & 74.8 & \underline{88.0} & 76.6 \\ 
    & \textbf{Fluency} \textcolor{red}{$\downarrow$} & 173.0 & \underline{161.5} & \textbf{116.5} & 276.0 & 174.5 & 42.4 \\
    & \textbf{GM} \textcolor{blue}{$\uparrow$} & 46.2 & \textbf{48.8} & 47.3 & 45.3  & \underline{48.2} & 40.9\\
    & \textbf{HM} \textcolor{blue}{$\uparrow$} & 37.6 & \underline{38.9} & \textbf{39.6} & 35.8 & 38.4 & 37.3\\ \hline
    \multirow{3}{*}{\textbf{DC}} & \textbf{Similarity} \textcolor{blue}{$\uparrow$} & \textbf{75.4} & \underline{73.1} & 66.5 & 59.5 & 52.0 & 36.6\\ 
    & \textbf{Strength} \textcolor{blue}{$\uparrow$} & 52.1 & \underline{63.5} & \textbf{67.3} & 55.9 & 43.9 & 55.5\\ 
    & \textbf{Fluency} \textcolor{red}{$\downarrow$} &  119.6 & 117.2 & \underline{91.4} & 172.6 & \textbf{69.5} & 36.4\\
    & \textbf{GM} \textcolor{blue}{$\uparrow$} & 43.5& \underline{46.0}& \textbf{46.3} & 40.1 & 37.8 & 38.4\\
    & \textbf{HM} \textcolor{blue}{$\uparrow$} & 37.4& \underline{38.9}& \textbf{40.0} & 34.8 & 35.5 & 36.9 \\ \hline
    \end{tabular}%
    }
    \end{scriptsize}
    \caption{Results for the YELP, IHS, and DIALOCONAN (DC) datasets. TTS stands for Two-time sampling. OS stands for One-Stage. PE stands for Prompt-Edit. D\&R stands for DeleteAndRetrieve. GM stands for geometric mean. HM stands for harmonic mean. Up-down arrows indicate whether higher or lower values are preferable. The best performance of each metric is highlighted in bold and the second-best is underlined. $^*$We do not compare the numbers of SSVAE with other methods due to its irrelevant (low similarity) outputs.}
    \label{tab:results-combined}
\end{table*}

\subsection{Real-World Datasets}
Tables \ref{tab:results-combined} present the results of various models on the three datasets. In a multi-stage setting, our two-time sampling (TTS) method outperforms the baseline model on all three datasets in HM and GM metrics, which comprehensively reflect overall performance across multiple metrics. TTS method demonstrates superior transfer strength on all three datasets, reaching 90.5\% on Yelp and exceeding other models by more than 7.4\% on Amazon and 30\% on IHS. Compared to PE, TTS achieves better fluency scores across all datasets. In terms of content preservation, it outperforms PE by over 14\%+ on both IHS and DC datasets. Compared to DR, TTS shows over 10\% improvement in similarity on the IHS dataset and over 14\% improvement on the DC dataset. On the Yelp dataset, TTS achieves comparable similarity while demonstrating better fluency scores. 

Our base method and one-stage method also outperform or are comparable to the baseline models on the HM and GM metrics, which validate the effectiveness of our generation framework. SSVAE’s significantly better fluency stems from its tendency to generate short, positive, and repetitive phrases. While these are favored by perplexity scores, they lead to mostly irrelevant outputs, as reflected in significantly lower similarity scores. This substantial gap between SSVAE and our methods underscores the importance of policy learning in addressing multi-stage decision problems effectively. We also have additional experiments on another style transfer dataset, a synthetic dataset, and a PPO algorithm shown in Appendix~\ref{app:more-rslt}. The results on other datasets are consistent with our findings here and the PPO algorithm cannot converge with limited training data and delayed rewards.

Overall, we observe that the two baseline methods designed for sentiment style transfer perform more closely to our methods on the Yelp dataset. However, their performance drops significantly on the new datasets. When addressing new, uncommon tasks with limited data, methods like PE, which rely on zero-shot or few-shot approaches, experience a significant performance decline. Similarly, retrieval-based methods like D\&R require a fair amount of training data as a retrieval database, leading to diminished performance in such scenarios. We acknowledge that there is considerable room for improvement in transfer strength for IHS and DC. This underscores that estimating the optimal natural language policy in a multi-stage setting remains a highly challenging task.

\subsection{Human Evaluation}
Table~\ref{tab:human-eval} presents the results of our human evaluation. Each sample's rating is the average of three raters' scores, and we report the overall average across all samples for each metric and model. The inter-rater agreement, measured by Cohen's kappa \cite{cohen1960coefficient}, is 0.37, indicating fair agreement. The Pearson correlation for all ratings is 0.65.

Our model's primary strength lies in its balanced performance across all three metrics: fluency, preservation, and transfer strength. This balance is particularly evident in the successful rewriting rate, which requires scores of 4 or higher on all three criteria. Here, our model significantly outperforms others. In comparison, DeleteAndRetrieve (D\&R) often fails to achieve successful rewrites due to substantial semantic shifts from the original input, while Prompt-Edit (PE) struggles with transfer strength. Specifically, our model's transfer strength surpasses PE by 1 point, and its preservation rate exceeds D\&R by 0.8 points.

\begin{table}[htbp]
\begin{tiny}
    \centering
    \resizebox{0.45\textwidth}{!}{
    \begin{tabular}{c|c|c|c|c}
    & Ours & PE & D\&R & Human \\ \hline
    Fluency & 3.8 & \textbf{4.3} & 3.8 & 4.9 \\ \hline
    Preservation & 4.1 & \textbf{4.5} & 3.3 & 4.7\\ \hline
    Sentiment & 4.0  & 3.0 & \textbf{4.2} & 4.3\\ \hline
    Suc. Rewrites & \textbf{45.3}\% & 41.0\% & 26\% & 78\%\\ \hline
    \end{tabular}}
    \caption{\small Results of human evaluations. The best performance of each criterion from all models except human references is highlighted in bold. All the criteria are the higher the better}
    \label{tab:human-eval}
    \end{tiny}
\end{table}
We noted that human evaluation trends for preservation (text similarity) and sentiment (transfer strength) generally align with automatic evaluations. However, there's a discrepancy in fluency assessments. PE's lower fluency score in automatic evaluations may be attributed to the frequent use of special characters (e.g., !!!) in its outputs, which automatic metrics may penalize but human raters don't necessarily perceive as disfluent.

\begin{table*}[ht]
\centering
\begin{small}
\begin{tabular}{c|l}
\hline
\textbf{Source 1} & \textcolor{red}{Avoid} reaching out for social support.\\
\hline
\textbf{Ours} & \textcolor{blue}{Consider} reaching out for social support.\\
\textbf{PE} & Reaching out for social \textcolor{orange}{good}.\\
\textbf{D\&R} & \textcolor{red}{Avoid} connecting with people.\\
\textbf{Human} & Reach out for social support.\\
\hline
\textbf{Source 2} & If I could give \textcolor{red}{less} stars, I would .\\
\hline
\textbf{Ours} & If I could give \textcolor{blue}{5} stars, I would.\\
\textbf{PE} & I could give \textcolor{blue}{15} stars, I would !!!\\
\textbf{D\&R} & \textcolor{orange}{Tender and full of fact that our preference menu is nice and full of flavor!}\\
\textbf{Human} & I would give an \textcolor{blue}{extra} star if it allowed me.\\
\hline

\textbf{Source 3} & definitely \textcolor{red}{disappointed} that I could \textcolor{red}{ not} use my birthday gift! \\
\hline
\textbf{Ours} & definitely \textcolor{blue}{delighted} that I can always use my Birthday present \textcolor{orange}{this year}.\\
\textbf{PE} & which definitely \textcolor{yellow}{ocked} that I could \textcolor{red}{not} use another birthday gift!\\
\textbf{D\&R} & \textcolor{orange}{thank you so much} for my birthday gift!\\
\textbf{Human} & definitely \textcolor{blue}{not disappointed} that I could use my birthday gift!\\
\hline

\textbf{Source 4} & I \textcolor{red}{didn't} even eat it . \\
\hline
\textbf{Ours} & I \textcolor{yellow}{didno't} even taste it...it was so \textcolor{blue}{good!}\\
\textbf{PE} & honestly, I \textcolor{red}{didn't} even immediately eat it .\\
\textbf{D\&R} & \textcolor{orange}{Tender and full of fact that our preference menu is nice and full of flavor!}\\
\textbf{Human} & I ate \textcolor{blue}{all} of it.\\
\hline
\end{tabular}
\end{small}
\caption{\small Four example sources and the corresponding outputs from different models. Red text represents negative signals, blue represents positive signals, orange represents hallucination generations, and yellow indicates grammatical issues.}
\label{tab:case}
\end{table*}

\subsection{Case Studies}

Table~\ref{tab:case} presents four typical outputs that demonstrate our model's capabilities across different scenarios. We highlight negative signals (e.g. words implying ineffective intervention) in red, positive signals in blue, hallucinations (information not present in the source) in orange, and grammatical issues in yellow. Sources 1, 3, and 4 are outputs from the second stage, while Source 2 is from the first stage. Source 1 from the IHS dataset illustrates our model's ability to perform basic editing operations, requiring only the replacement of ``Avoid'' to shift the effectiveness of the intervention. In contrast, the Prompt-Edit (PE) model, while achieving a relatively positive tone, introduced irrelevant content (``social good''), leading to lower similarity scores. The DeleteAndRetrieve (D\&R) model failed to alter the intervention at all.

Sources 2 and 3 demonstrate our model's proficiency in substitution operations. In Source 3, although our model produced some hallucination, it effectively preserved the overall content and sentiment. Source 4 presents a more challenging case where all models, including ours, struggle to produce a perfect output as it may require contextual understanding. Notably, the D\&R model generated identical outputs for both Source 2 and Source 4, suggesting limitations in its retrieval method. Our model, despite introducing a grammatical issue (specifically, a misspelling of ``didn't'' as ``didno't''), successfully inserted positive signals and flipped the overall sentiment of the sentence.

It's important to note that our model achieves these varied transformations without explicitly restricting gradient ascent to basic editing operations like insertion, deletion, and substitution. Instead, it learns to perform these operations naturally across different stages of the decision-making process. This flexibility demonstrates the effectiveness of our framework in addressing multi-stage decision problems with natural language actions, adapting to various transformation requirements without hard-coded rules. These case studies underscore the versatility of our approach in handling diverse scenarios in sentiment transfer tasks, from simple negation removal to more complex semantic transformations, while also revealing areas for potential improvements, such as reducing hallucinations and maintaining grammatical correctness in challenging cases.

\section{Discussion and Conclusion}
We introduced a causal framework for multi-stage decision-making that addresses the challenges inherent in continuous, high-dimensional natural language action spaces. While recent reinforcement learning approaches like PPO can handle delayed rewards in language tasks, they typically require learning multiple models and substantial training data. Our Q-learning based approach offers a more data-efficient alternative, optimizing policies through gradient ascent on language embeddings using a single model. We demonstrated this efficiency through significant improvements over competitive baselines across multiple metrics and datasets. Human evaluations corroborate these findings, showing more balanced performance across transfer strength, content preservation, and fluency. While our results on mental health interventions, hate speech countering, and sentiment transfer are promising, substantial challenges remain. Future work could address performance gaps on novel tasks and explore scaling to more complex action spaces. By providing a data-efficient framework for learning optimal policies in natural language spaces, our work establishes a foundation for developing more effective AI systems for sequential text refinement tasks.

\section*{Acknowledgments} This work was supported in part by the Office of Naval Research under grant number N00014-23-1-2590, the National Science Foundation under Grant No. 2231174, No. 2310831, No. 2428059, No. 2435696, No. 2440954, and a Michigan Institute for Data Science Propelling Original Data Science (PODS) grant.

\bibliographystyle{apalike} 
\bibliography{example_paper}

\newpage
\appendix
\section{Prompts to build IHS dataset}
There are 9 questions asked in the PHQ-9 questionnaires:
\begin{itemize}
    \item PHQ-1 Little interest or pleasure in doing things
\item PHQ-2 Feeling down, depressed or hopeless
\item PHQ-3 Trouble falling asleep, staying asleep or sleeping too much
\item PHQ-4 Feeling tired or having little energy
\item PHQ-5 Poor appetite or overeating
\item PHQ-6 Feeling badly about yourself, or that you are a failure, or that you have let yourself or your family down
\item PHQ-7 Trouble concentrating on things such as reading the newspaper or watching TV
\item PHQ-8 Moving or speaking so slow that others could have noticed or the opposite, being so fidgety or restless that you have been moving around a lot more than usual
\item PHQ-9 Thoughts that you would be better off dead or hurting yourself in some way
\end{itemize}

We use gpt-4o-2024-08-06 to generate medical interventions for each criterion. The prompt is ``If someone is having [PHQ-x], what should they and shouldn't do? You should list 25 short suggestions starting with ``you should'' and 25 with ``you shouldn't''.'' where [PHQ-x] is one of the conditions above. When building the IHS dataset, for each patient’s most deteriorated condition, there is a 50\% probability of receiving a medical intervention randomly selected from the list of 25 “you should” suggestions, while the other 50\% comes from “you shouldn’t” suggestions. We will then remove ``You should'' and ``You shouldn't'' in the selected suggestions so that the selected suggestions become either effective or ineffective interventions.
\section{Prompts to build synthetic dataset}\label{app:syn-prompt}
We use gpt-4o-2024-08-06 to generate synthetic datasets. The prompt is ``Please add the word `[WORD]' to the appropriate position of the given sentence without changing other words. Your output should directly output the revised sentence without any additional text. The sentence to be modified is: [TEXT]''
\section{Experiment setup}\label{app:exp-set}
\subsection{T5 and \textit{Repeat} training}
As mentioned above, we tuned a T5-base model~\cite{raffel2020exploring} following the \textit{Repeat} prompt. We randomly selected 10,000 summaries from the CNN/DailyMail dataset~\cite{hermann2015teaching}. The training input is ``Repeat: " followed by the summary, while the target output is the summary itself. The learning rate is 5e-3, the batch size is 16, and the number of training epochs is 2. The average rouge score between the outputs and original summaries on another 2000 randomly selected test summaries is 0.99, which validates the repeat framework. 
\subsection{Baseline Models}
\subsubsection{SSVAE}
Following~\cite{zhang2024causal}, we implemented a baseline without Q-learning, omitting maximization and outcome propagation. The architecture of SSVAE~\cite{kingma2014semi} is similar to our main model. The low-dimensional embedding reduced from the concatenated token embeddings of the inputs is used to predict the tone of the input texts. The classifier is still a three-layer transformer encoder. Then an one-hot vector of the predicted label is then concatenated with the low-dimensional embedding and fed into the decoder to reconstruct the original embedding during the training of SSVAE. Thus, the loss for SSVAE is the sum of the reconstruction loss, task classification loss, and KL divergence loss. In the test time, we aim to change the tone of the input to positive. Therefore, the one-hot vector concatenated with the low-dimensional embedding always represents the positive class.
\subsubsection{DeleteAndRetrieve}
The D\&R~\cite{li2018delete} method identifies and deletes attribute-specific words or phrases (e.g. negative signals) from the source sentence to extract the core content. It retrieves a sentence from the target attribute corpus that has similar content to the extracted core. Then it uses a neural sequence-to-sequence model to generate the output by combining the content words from the source sentence with attribute markers from the retrieved target sentence. This approach allows for explicit separation of content and attribute and outperformed previous adversarial methods in human evaluations across multiple datasets. This framework can be easily adapted to the two non-style transfer datasets as long as we can define a corpus with positive outcomes and a corpus with negative outcomes.
\subsection{Prompt-Based Editing}
~\citet{luo-etal-2023-prompt} presents a new approach to text style transfer using prompt-based editing. It uses language models to classify sentence style, then performs discrete word-level edits using steepest-ascent hill climbing to maximize a scoring function combining style, fluency, and semantic similarity. Evaluated on sentiment and formality transfer tasks across three datasets, this method outperforms existing prompting systems, including those using much larger models. The approach avoids error accumulation issues in autoregressive generation and allows for more controlled style transfer. This prompting system can be effectively adapted to the two non-style transfer datasets. When addressing style transfer, the original prompt is: ``The sentiment of the text \{text\} is: '', and the model compares the probabilities of the next word being ``positive'' or ``negative''. For DIALOCONAN, the prompt is: “The tone of the text \{text\} is: '', and the model compares the probabilities of the next word being ``aggressive'' or ``respectful''. For IHS, the prompt is: ``The health intervention {text} is: '', and the model compares the probabilities of the next word being ``effective'' or ``ineffective''.
\subsection{Text Classifiers}
The text classifier for each stage is a transformer encoder~\cite{vaswani2017attention}. The model's input is the text embedding representation from the T5 encoder, so the input size is 768. The hidden size is 128, the number of heads is 8, the number of encoder layers is 3 for Yelp and 8 for Amazon and the dropout rate is 0.1. The learning rate is 1e-4, the batch size is 16 and the number of epochs is 15 for all stages.
\subsection{Gradient Ascent}
The number of iterations of gradient ascent is 10 for the second stage and 15 for the first stage. When decoding, we adopt a beach search with a beam size of 3. The maximum output length is 256.
\section{More results}\label{app:more-rslt}
\subsection{Amazon Dataset}
We also test our methods on Amazon Review~\cite{he2016ups} dataset. Similar to the YELP review dataset, it contains over 200k sentiment-labeled reviews. Our model attempts to transform sequences of texts from negative to positive sentiment. The training data includes 900 pairs for each combination of the outcomes. The test set includes 500 negative sentences. The training data is non-parallel and lacks references. The results are consistent with our findings above.
\begin{table*}[htbp]
    \centering
    \begin{small}
    \resizebox{\textwidth}{!}{%
    \begin{tabular}{cc|c|c|c|c|c|c}
    \hline
    & \textbf{} & \textbf{Ours} & \textbf{Ours (two-time sampling)} & \textbf{Ours (single-stage)} & \textbf{PE} & \textbf{D\&R} & \textbf{SSVAE}$^*$ \\ \hline
    
    \multirow{3}{*}{\textbf{Amazon}} & \textbf{Similarity} \textcolor{blue}{$\uparrow$} & 53.8 & 54.4 & 62.5 & \underline{62.8} & \textbf{72.7} & 17.6 \\ 
    & \textbf{Strength} \textcolor{blue}{$\uparrow$} & 54.5 & \textbf{58.4} & 48.2 & 41.6 & 40.2 & \underline{54.8} \\ 
    & \textbf{Fluency} \textcolor{red}{$\downarrow$}& \underline{204.0} & 235.9 & \textbf{130.2} & 247.4 & 219.1 & 40.9 \\ 
     & \textbf{GM} \textcolor{blue}{$\uparrow$} & 38.1 & \underline{38.7} & \textbf{39.6} & 36.2 & 37.9 & 29.6 \\ 
    & \textbf{HM} \textcolor{blue}{$\uparrow$} & \underline{33.3} & \underline{33.3} & \textbf{35.1} & 31.6 & 32.4 & 26.7 \\ \hline
    \end{tabular}%
    }
   
    \caption{\small Results for the Amazon datasets. PE stands for Prompt-Edit and D\&R stands for DeleteAndRetrieve. GM stands for geometric mean and HM stands for harmonic mean. Up-down arrows indicate whether higher or lower values are preferable. The best performance of each metric is highlighted in bold and the second-best is underlined. $*$ indicates that we did not compare the numbers of SSVAE with other methods due to its low generation quality.}
    \label{tab:results-amazon}
    \end{small}
\end{table*}

\subsection{Synthetic Datasets}
We constructed two synthetic settings to validate our model, using pairs of keywords as trigger signals for the outcome. In the first setting, sentences containing \textit{good} are labeled positive, while those with \textit{bad} are negative. We randomly selected 2,000 reviews from the Yelp dataset without these words and used ChatGPT to insert \textit{good} or \textit{bad} into 1,000 reviews each, ensuring sentence fluency. We then created 250 two-sentence sequences for each label pair, with the final outcome based on the number of positive signals. The second setting added another signal pair: \textit{sad} and \textit{happy}, randomly replacing half of the existing signals (e.g., changing \textit{good} to \textit{happy}) while maintaining sentiment. This tests our model's ability to rewrite sentences to include positive signals with different numbers of possible signal pairs. In the first setting, each output should contain \textit{good}, while in the second, it should include one of {\textit{good}, \textit{happy}}. For synthetic datasets, we use the embedding from the final gradient ascent iteration as $a^*_t$, without NLL-based selection. We evaluate using 5-fold cross-validation.

Table~\ref{tab:syn-results} presents the synthetic results, showing the accuracy of transferring negative signals to positive ones. Our model demonstrates high efficacy with a single pair of signals, achieving 88.0\% and 96.9\% accuracy for the second and first stages, respectively. With two signal pairs, the accuracy decreases to 80.5\% and 74.1\% for the second and first stages. The higher first-stage accuracy may be attributed to naturally shorter inputs, but it's important to note that second-stage performance directly influences first-stage results due to the derivation of first-stage pseudo outcomes from the second stage. These results highlight our model's capability in handling varying levels of signal complexity, while also revealing the interdependence of performance across stages in multi-stage decision-making tasks.

\begin{table}[htbp]
\begin{tiny}
    \centering
    \resizebox{0.35\textwidth}{!}{
    \begin{tabular}{c|c|c}
    Setting & One Pair & Two Pairs \\ \hline
    Stage 2  &  88.0    & 80.5  \\ \hline
    Stage 1  &  96.9    & 74.1   \\ \hline
    \end{tabular}}
    \caption{\small Synthetic data performance. The numbers represent the accuracy of converting negative signals into positive signals.}
    \label{tab:syn-results}
    \end{tiny}
\end{table}

The presence of multiple signals increases the task’s complexity. Real-world data resembles a multi-signal scenario, where different inputs may contain different signals. Our model needs to identify and rewrite these signals accordingly. In the two-pair setting, if we count whether the negative signal was deleted (not necessarily rewritten into a positive signal), the accuracy for the first and second stages reaches 88.7\% and 97.0\%, respectively. This is beneficial for real-world data, as in such cases, we do not always need to rewrite into a positive signal to achieve outcome transfer; often, simply deleting the negative signal can be sufficient. Overall, our model maintains a relatively high transfer rate on the synthetic dataset, demonstrating its potential for multi-stage decision-making problems with natural language treatments.

\begin{figure}[h]
\begin{small}
    \centering
    \includegraphics[width=.5\textwidth]{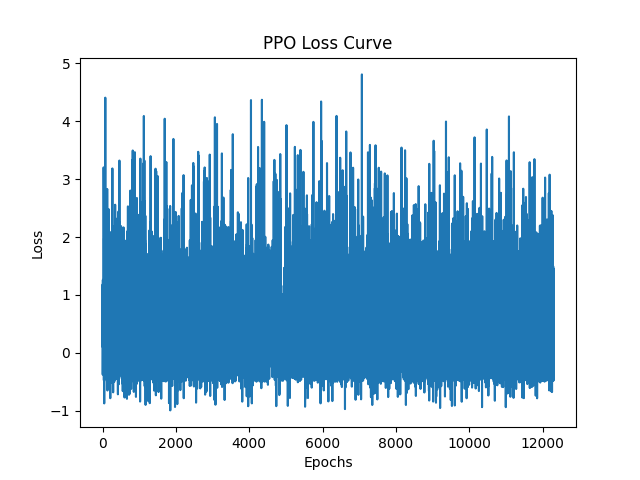}
    \caption{The loss curve of PPO training on the DIALOCONAN datasets. The algorithm cannot converge with limited training data and delayed rewards.}
    \label{fig:ppo-loss}
    \end{small}
\end{figure}
\subsection{PPO baseline}\label{app:ppo}
We fine-tune a GPT-2 model~\cite{radford2019language} via PPO for text refinement in each task. The reward is defined as the outcome gain, computed as the difference between the outcomes of the new and old trajectories, using a RoBERTa-based classifier~\cite{liu2019roberta} fine-tuned for each task. The value function, estimated by another tuned RoBERTa classifier, predicts the expected outcome of a state. The training follows standard PPO updates with a learning rate of $3 \times 10^{-5}$ and a discount factor of 0.99. The clipping parameter is set to 0.2. The model is trained for five epochs, with each epoch sampling a number of trajectories equivalent to five times the dataset size. As shown in Figure~\ref{fig:ppo-loss}, the model does not converge due to limited training data and the challenges of learning from delayed rewards.

\section{Questionaire for Human Evaluation}\label{app:human-eval}
Human raters will be asked three questions when given an input sample and a rewritten output from a model:

1. On a scale of 1 to 5, how well does the rewritten text preserve the content of the original?

2. On a scale of 1 to 5, how positive is the sentiment of the rewritten text?

3. On a scale of 1 to 5, how fluent is the rewritten text?

\end{document}